\documentclass[review]{elsarticle}

\usepackage{lineno,hyperref}
\modulolinenumbers[5]

\usepackage{graphicx}
\usepackage{epstopdf}
\usepackage{amsmath}
\usepackage{amsfonts}
\usepackage{pdflscape}
\usepackage{rotating}

\journal{Any Journal}

\begin{document}

\begin{frontmatter}

\title{Monotonic classification: an overview on algorithms, performance measures and data sets}

\author{Jos\'e-Ram\'on Cano}
\address{Dept. of Computer Science, University of Ja\'en, EPS of Linares, Avenida de la Universidad S/N, Linares 23700, Ja\'en, Spain.}
\ead{jrcano@ujaen.es}

\author{Pedro Antonio Guti\'errez}
\address{Department of Computer Science and Numerical Analysis, University of C\'ordoba, C\'ordoba, Spain.}
\ead{pagutierrez@uco.es}

\author{Bartosz Krawczyk}
\address{Department of Computer Science, Virginia Commonwealth University, Richmond, VA, 23284, USA.}
\ead{bkrawczyk@vcu.edu}

\author{Micha\l{} Wo\'{z}niak}
\address{Department of Computer Science, Wroc\l{}aw University of Technology, Wyb. Wyspia\'{n}skiego 27, 50-370 Wroc\l{}aw, Poland.}
\ead{michal.wozniak@pwr.edu.pl}

\author{Salvador Garc\'ia\corref{cor1}}
\address{Department of Computer Science and Artificial Intelligence, University of Granada, 18071, Granada, Spain.}
\ead{salvagl@decsai.ugr.es}
\cortext[cor1]{Corresponding author}

\begin{abstract}
Currently, knowledge discovery in databases is an essential step to identify valid, novel and useful patterns for decision making. There are many real-world scenarios, such as bankruptcy prediction, option pricing or medical diagnosis, where the classification models to be learned need to fulfill restrictions of monotonicity (i.e. the target class label should not decrease when input attributes values increase). For instance, it is rational to assume that a higher debt ratio of a company should never result in a lower level of bankruptcy risk. Consequently, there is a growing interest from the data mining research community concerning  monotonic predictive models. This paper aims to present an overview about the literature in the field, analyzing existing techniques and proposing a taxonomy of the algorithms based on the type of model generated. For each method, we review the quality metrics considered in the evaluation and the different data sets and monotonic problems used in the analysis. In this way, this paper serves as an overview of the research about monotonic classification in specialized literature and can be used as a functional guide of the field.
\end{abstract}


\end{frontmatter}

\section{Introduction}
\label{intro}

Data mining, as a key stage in the discovery of knowledge, is aimed at extracting models that represent data in ways we may not have previously taken into consideration \cite{Witten2011}. Among all the data mining alternatives, we focus our attention on classification as a predictive task \cite{Saleh2017,Tama2017}. There is a particular case of predictive classification where the target class takes values in a set of ordered categories. In that case we are referring to ordinal classification or regression \cite{Gutierrez2016}. In addition, the classification task is defined as monotonic classification in those cases where we have ordered domains of attributes and a monotonic relationship between an evaluation of an object on the attributes and its class assignment \cite{kotlowski13}.

Monotonicity is a type of background knowledge of vital importance for many real problems, which is needed to obtain more accurate, robust and fairer models of the data considered. In this way, monotonicity can be found in different environments such as economics, natural language or game theory \cite{kotlowski13}, as well as the evaluation of courses at teaching institutions \cite{cano2017prototype}.

Some important examples of real problems where this kind of background knowledge has to be considered are now analyzed. For bankruptcy prediction in companies \cite{kim2003discovery}, the appropriate actions should be taken in time but considering the information based on financial indicators taken from their annual reports. The monotonicity is present in the comparison of two companies where one dominates the other for all financial indicators. Because of this dominance, the overall evaluation of the second one should not be higher than the that of the first. In this way, monotonic classification has been applied to predict the credit rating score used by banks \cite{chen2014credit}. Another example is the house pricing problem \cite{potharst02}, in which we should assure that the price of a house increases with an increase of the number of rooms or with the availability of air conditioning, and that it decreases with, for example, the pollution concentration in the area.

Considering monotonicity constraints in a learning task is motivated by two main facts\cite{ben-david92}: (1) the size of the hypothesis space is reduce, what facilitates the learning process; (2) other metrics besides accuracy, such as the consistency with respect to these constraints, can be used by experts to accept or reject certain models. 

In this way, the need of handling background knowledge about ordinal evaluations and monotonicity constraints in the learning process has led to the development of new algorithms. The interest in the field of monotonic classification has significantly increased  \cite{gutierrez2016current,zhu2017monotonic}, leading to a growing number of techniques and methods. Apart from these algorithmic developments, different quality measures have been presented to measure the consistency with respect monotonicity constraints.

Given that, up to the authors' knowledge, there are no functional guides for this domain of study, it can be difficult to obtain a general overview of the state of the art. Because of this reason, this paper presents an overview on the monotonic classification field, including:

\begin{itemize}
	\item A systematic review of the techniques proposed in the literature.
	\item A taxonomy to categorize all the existing algorithms, including whether or not there is publicly available software related to them.
	\item The quality measures applied to evaluate the performance of monotonic classifiers in the literature. These metrics analyze the performance both in terms of accuracy and degree of fulfillment of the monotonicity constraints.
	\item Finally, the data sets considered in every proposal and a summary of which are the most used and where they can be found. 
\end{itemize}

The remainder of this paper is structured as follows. Section \ref{sec:background} presents a definition of the monotonic classification problem. Section \ref{sec:taxonomy} shows an overview of the monotonic methods and the taxonomy proposed to categorize them. Section \ref{subsec:qualityindexes} offers an analysis of the quality metrics considered in monotonic classification. 
Section \ref{sec:datasets} presents the data sets evaluated in the literature, highlighting the most popular ones and where they can be found. Finally, Section \ref{sec:conclusions} is devoted to the conclusions reached.

\section{Definition of monotonic classification}
\label{sec:background}

\begin{figure*}
	\includegraphics[width=0.95\textwidth]{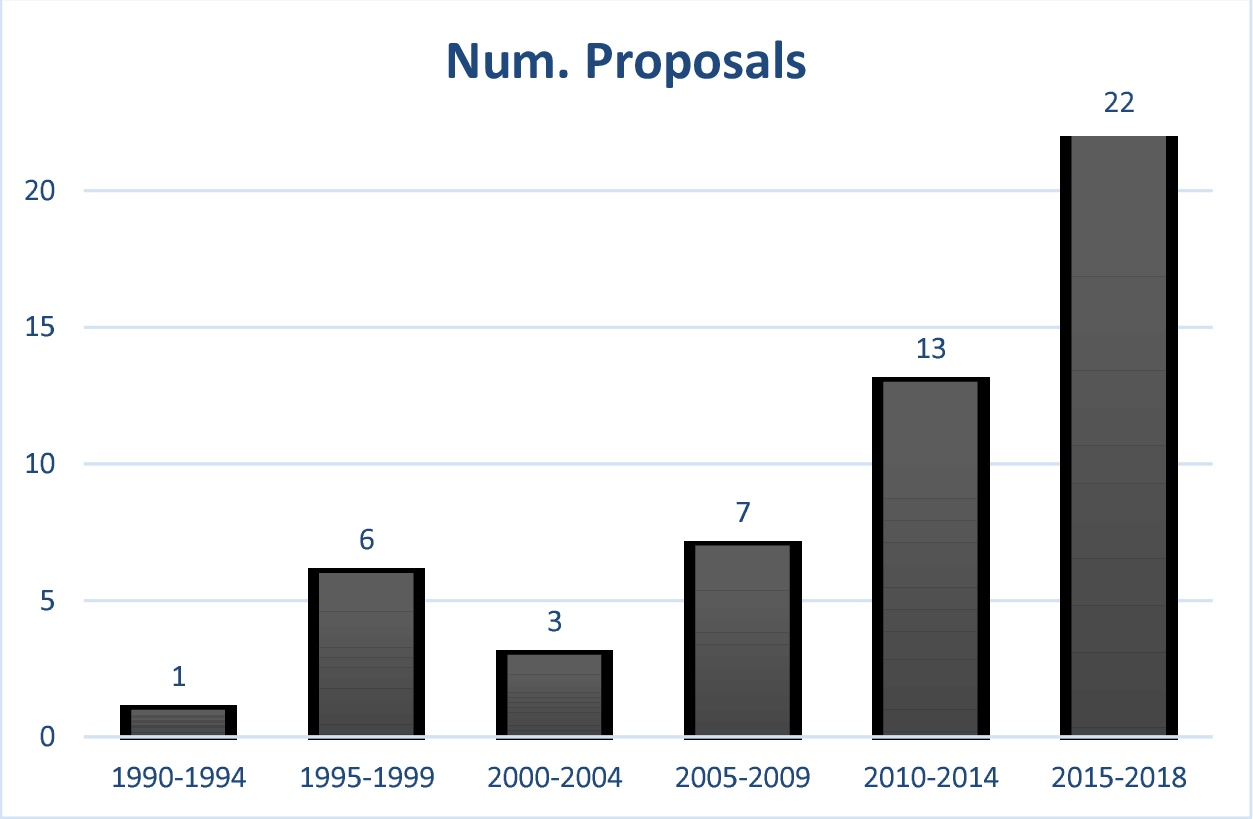}
	\caption{Number of monotonic classification proposals over time.}
	\label{fig:numProposals}       
\end{figure*}

The process of data knowledge discovery in databases is a key objective for organizations to make accurate and timely decisions and recognize the value in data sources. One of the main stages within the process is data mining \cite{Witten2011}, where models are extracted from the input data collected. These models are used to support people in making decisions about problems that may be rapidly changing and not easily specified in advance (i.e. unstructured and semi-structured decision problems). Among all kinds of models, we focus our attention on classification algorithms, where the goal is to predict the value of a target variable. When the target variable exhibits a natural ordering, we are talking about ordinal classification (also known as ordinal regression) \cite{Cardoso2011,gutierrez2016current,Gutierrez2016,KotlowskiThesis08}. The order of the categories can be exploited to construct more accurate models in those application domains involving preferences, like social choice, multiple criteria decision making, or decision under risk and uncertainty. For example, in a factory a worker can be evaluated as ``excellent'', ``good'' or ``bad'', or a credit risk can be rated as ``AAA'', ``AA'', ``A'' or ``A-''. A particular case of ordinal classification is monotonic classification \cite{gutierrez2016current}. The interest in monotonic classification of the scientific community has increased in the last years. This  fact can be corroborated in Figure \ref{fig:numProposals}, where the number of proposals in the specialized literature is represented over time.

Classification problems where there exist a background knowledge in the form of ordinal evaluations and monotonicity constraints are very common. In this kind of problems, the order properties of the input space are exploited, by using the available knowledge in terms of dominance relation (one sample dominates another when each coordinate of the former is not smaller than the respective coordinate of the latter). Monotonicity constraints require that the class label assigned to a pattern should be greater or equal than the class labels assigned to the patterns it dominates. As an example, consider a monotonicity constraint relating one input attribute and the target class. In this case, a sample in the data set with a higher value of the input attribute should not be associated to a lower class value, as long as the other attributes of the sample are fixed. A monotonicity constraint always involves one input attribute and the class attribute, and there should be, at least, one monotonicity constraint (to distinguish monotonic classification from ordinal regression). Monotonicity constraints can be either direct (as the example presented before) or inverse (if the value of the attribute decreases, the class value should not increase). Usually, in real monotonic classification problems, the monotonicity constraints are assumed only for a subset of the input features.

As a descriptive example, we can consider student evaluation in a college, the students being evaluated with a rating between 0 and 10. We consider three students (Student A, B and C) with 22 evaluations each one and a final mark. We consider that all the input attributes (22 evaluations) have a direct monotonic assumption with respect to the output value (final qualification, represented in bold face):
\begin{itemize}
	\item Student A: 5,5,5,5,7,6,5,5,5,5,5,5,6,5,5,6,6,6,5,5,5,5,\textbf{4}.
	\item Student B: 3,5,3,4,7,3,3,5,3,3,3,3,6,3,3,4,3,6,4,3,5,3,\textbf{5}.
	\item Student C: 2,2,1,2,1,2,2,3,2,2,1,2,3,2,2,3,3,2,2,1,2,3,\textbf{2}.
\end{itemize}
As can be observed, there is a monotonic violation involving two samples (Student A and B), where Student B, who has worse or equal evaluation marks than Student A, presents a higher final qualification. On the other hand, there are no monotonic violations when considering Student C with respect to both Students A and B.

Now, we formally define a classification dataset with ordinal labels and monotonicity constraints. Let assume that patterns are described using a total of $f$ input variables with ordered domains, $\mathbf{x}_i\subseteq \mathbb{R}^f$, and a class label, $y_i$, from a finite set of $C$ ordered labels, $y_i\in \mathcal{Y}=\{1,...,C\}$. In this way, the data set $D$ consists of \textit{n} samples or instances $D=\{(\mathbf{x}_1,y_1),...,(\mathbf{x}_n,y_n)\}$. As previously discussed, a \textit{dominance relation}, $\succeq$, is defined as follows:
\begin{equation}
\mathbf{x}\succeq \mathbf{x}'\Leftrightarrow x^s \ge x^{s'} \forall s\text{ with a monotonicity constraint},
\end{equation}
where $x^s$ and $x^{s'}$ are the $s$-th coordinates of patterns $\mathbf{x}$ and $\mathbf{x}'$, respectively. In other words, $\mathbf{x}$ dominates $\mathbf{x}$' if each coordinate of $\mathbf{x}$ is not smaller than the respective coordinate of $\mathbf{x}'$.

Samples $\mathbf{x}$ and $\mathbf{x}'$ in space \textit{D} are \textit{comparable} if either $\mathbf{x}' \succeq \mathbf{x}'$ or $\mathbf{x}' \succeq \mathbf{x}$. Both $\mathbf{x}$ and $\mathbf{x}'$ are
\textit{incomparable} otherwise. Two examples $\mathbf{x}$ and $\mathbf{x}'$ are \textit{identical} if ${x}^j = x^{j'}, \forall j\in\{1,\ldots,f\}$, and they are \textit{non-identical} if $\exists j$ for which $x^j \neq x^{j'}$.

A pair of comparable examples $(\mathbf{x},y)$ and $(\mathbf{x'},y')$ is said to be monotone if \footnote{Recall that $y,y'\in \mathcal{Y}=\{1,...,C\}$, so that every two labels can be compared using the ordinal scale}:
\begin{equation}
\mathbf{x}\succeq \mathbf{x}' \wedge \mathbf{x} \neq \mathbf{x}'  \wedge y\geq y',
\end{equation}
or
\begin{equation}
\mathbf{x} = \mathbf{x}' \wedge y=y'.
\end{equation}

A data set $D$ with $n$ examples is monotone if all possible pairs of examples are either monotone or incomparable. It is worth mentioning that the previous notation was expressed for direct monotonicity constraints, but it could be changed to consider inverse ones.

\section{A taxonomy for monotonic classification algorithms}
\label{sec:taxonomy}

This section presents and describes the proposals in the specialized literature for monotonic classification, deriving a taxonomy about them.
The categorization is based on the goal of the different methods, the heuristics followed and the models generated by each algorithm. In this sense, the algorithms proposed can be divided into:

\begin{enumerate}
	\item Monotonic Classifiers, aiming at the generation of predictive models satisfying the monotonicity constraints partially or totally. There are several families of classifiers depending on the type of model they build:

	\begin{itemize}
		
		\item Instance based classifiers. These algorithms do not build a model but they directly use the instances of the data set of to make classification decisions.
		
		\item  Decision trees or classification rules. In this case, the models built involve readable production rules in forms of decision trees or set of rules.
		
		\item Ensembles \cite{rokach2010ensemble} or multiclassifiers. This group is composed by methods which use several classifiers to obtain different responses, which are aggregated into a global classification decision. Two classical approaches are considered:
		
		\begin{itemize}
			\item Boosting: a number of weak learners are combined to create a strong classifier able to achieve accurate predictions. These algorithms use all data to train each learner, but the instances are associated with different weights representing their relevance in the learning process. If an instance is misclassified by a weak learner, its weight is increased so that subsequent learners give more focus to them. This process is applied iteratively.
			\item Bagging: it chooses random subsets of samples with replacement of the  data set, and a (potentially) weak learner is trained from each subset.
		\end{itemize}
		
		\item Neural Networks. These are biologically inspired models, where the function relating inputs and target attribute consists of a set of building blocks (neurons), which are organized in layers and interconnected. An iterative training process is performed to obtain the values of connection weights.
		
		\item Support Vector Machines. This family considers support vector machines based learning and derivatives.

		\item Hybrid. This last set of algorithms considers the combination of different classification algorithms into a hybrid one (for example, rule and instance-based learning).
		
		\item Fuzzy Integral. These algorithms are based on the use of the Choquet integral which can be seen as a generalization of the standard (Lebesque) integral to the case of non-additive measures \cite{tehrani2012learning}.
		
	\end{itemize}

	\item Monotonic Preprocessing  refines the data sets in order to improve the performance of monotonic classification algorithms:

	\begin{itemize}
		
		\item Relabeling. These methods change the label of the instances to minimize the number of monotonicity violations present in the data set.
		
		\item Feature selection. Their objective is to obtain the most relevant features to improve monotonic classification performance. 
		
		\item Instance selection. In this case, a subset of samples is selected from the data set with the objective of deriving better monotonic classifiers. 
		
		\item Training Set Selection. The heuristic followed by this set of algorithms must be generic in such a way that the selected set is the one that reports the highest performance regardless of the classifier subsequently used.
		
	\end{itemize}

\end{enumerate}

\begin{figure*}
	\includegraphics[width=1.0\textwidth]{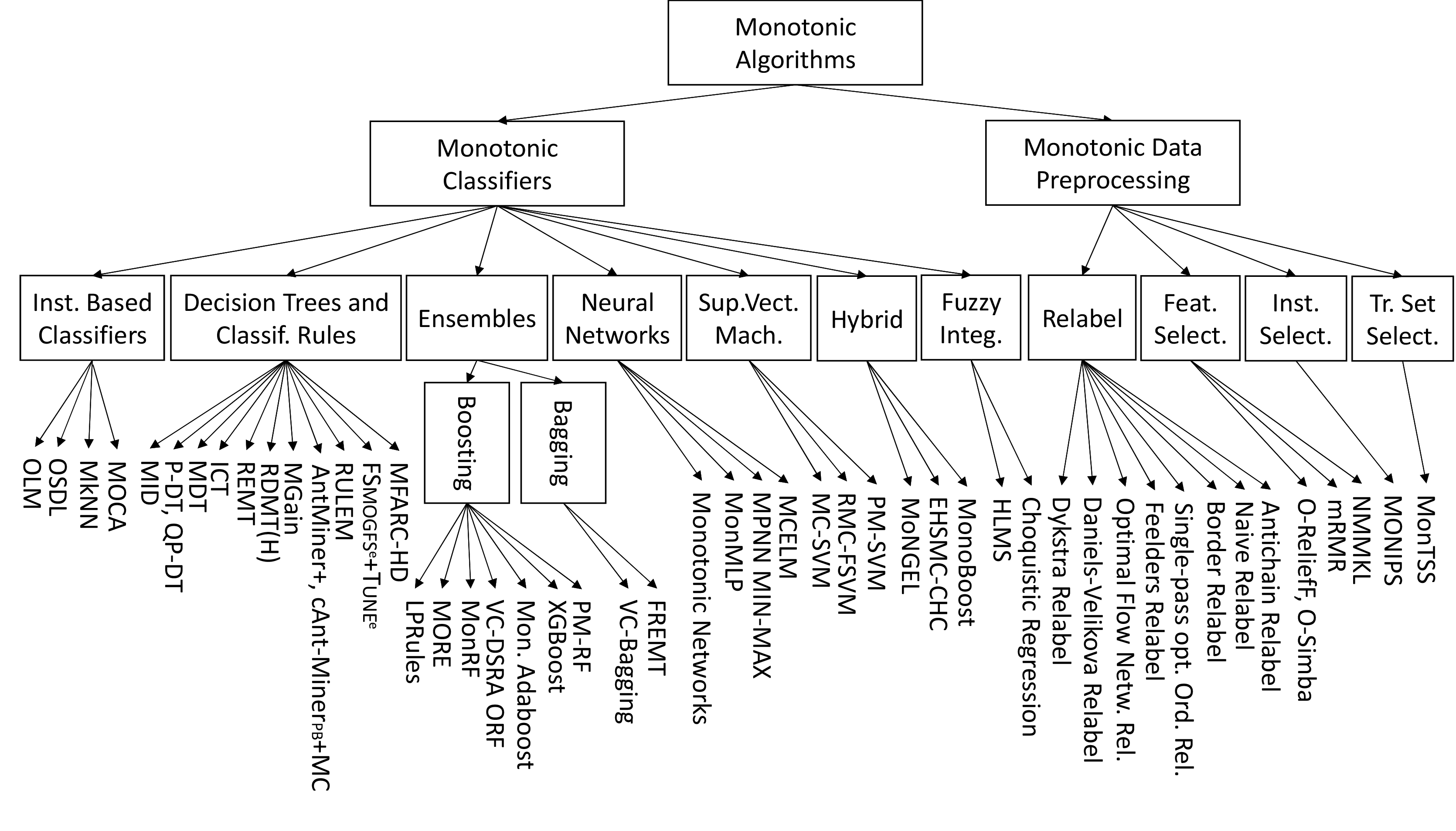}
	\caption{Monotonic algorithms taxonomy.}
	\label{fig:taxonomy}       
\end{figure*}

Figure \ref{fig:taxonomy} shows the proposed taxonomy and Table \ref{tab:methods} the summary of all the monotonic classifiers found in the specialized literature. The first column of the table contains the year of the proposal, the second is the reference and the third is the proposal name. We also show in the fourth and fifth columns, whether or not the algorithm  requires a total monotonic input data set and whether or not it produces complete monotonic output models, respectively. The fourth column is referred by some authors as dealing with partial monotonic data sets, i.e. \cite{daniels10}. Seventh and eighth columns present the non monotonic classification algorithms used as baseline to compare the method and the monotonic classifiers used for comparison in the experimental analysis conducted in each paper. The last column offers whether or not the algorithm's source code is publicly available and, if it is, the name of framework in which we can find it.

Next, we provide a description of the methods in each family.

%
%

\begin{landscape}
	
	\begin{table}
		\caption{Monotonic classification methods reviewed}
		\label{tab:methods}       
		
		\centering
\tiny
\begin{tabular}{llcccccl}
	\hline\noalign{\smallskip}
	&           &      &    Require & Completely &\multicolumn{2}{c}{Comparison versus} &  \\
	\cline{7-8}
	Year & Reference & Abbr. name &  Input Monot. & Monot. Output & Classical methods  & Monotonic methods &  Code available in \\
	\noalign{\smallskip}\hline\noalign{\smallskip}
	
	1992 &  \cite{ben-david92} & OLM &  Yes & Yes &  C4, ID3 & None  & \cite{frank2016weka} in WEKA \\		
	
	1995 & \cite{ben-david95} & MID &  No  & No  &  ID3 & OLM &  Not Available \\ 
	
	1995 & \cite{grabisch1995new}  & HLMS  & No & Yes & None & None & Not Available\\
	
	1997 & \cite{sill1997monotonic} & Monotonic Networks  & Yes  & Yes & None  & None & Not Available \\
	
	1999 & \cite{MAKINO1999}	 & P-DT R, QP-DT R & Yes, No  & Yes, No  & ID3  & MID  & Not Available \\

	2001 &  \cite{dembczynski09}  & MORE & Yes & Yes & SVM, J48, kNN  & None & Not Available \\

	2003 & \cite{lee2003monotonic}	&  MDT & No  & Yes  & CART & None  & Not Available\\

	2005 & \cite{lang2005monotonic} & MonMLP & Yes &  Yes  &	None & 	None & in CRAN\\
	
	2008 & \cite{lievens08} &  OSDL & Yes & Yes &  None  & None  
	& \cite{frank2016weka} in WEKA \\

	2008 & \cite{duivesteijn08mKNN} & MkNN & Yes & Yes &  kNN & None  & Not Available \\ 

	2008 & \cite{barile2008nonparametric} & MOCA  & Yes   & Yes & OSDL  & None   & Not Available \\ 
	
	2009 & \cite{van2009isotonic} & ICT & No & No & None & None & Not Available \\
	
	2009 & \cite{kotlowski09}  &   LPRules  & Yes & Yes &  J48, SVM & OLM, ICT &  Not Available \\
	
	2010 & \cite{daniels10} & MPNN MIN-MAX & No & No& None & None & Not Available\\
	
	2010 & \cite{Blaszczynski2010} & VC-bagging  & No & No  & None & OLM, OSDL &  Not Available\\		
	
	2012 & \cite{hu2012rank} & REMT & No & No & CART, Rank Tree  & OLM, OSDL  & Not Available \\
	
	2012 & \cite{tehrani2012learning} & Choquistic Regression &  Yes & Yes &  MORE & LMT, Logistic Regression  & Not Available \\
	
	2014 & \cite{chen2014credit} & MC-SVM & Yes & Yes  & SVM & None & Not Available\\
	
	2015 &  \cite{zhang2015induction}	 & MGain & No & No  & C4.5 & None & Not Available \\
	2015 &  \cite{qian2015fusing} & FREMT & No & No & None & REMT &  Not Available \\	
	2015 &  \cite{Gonzalez2015} & MonRF & No & No & None & OLM, OSDL, MID  & Not Available\\	
	
	2015 & \cite{wang2015ordinal} & VC-DRSA ORF & No & No  & None & None & \cite{blaszczynski2013jmaf} in jMAF \\	
	
	2015 &	\cite{marsala2015rank} 	 & RDMT(H) & No & No  & None & MID, ICT  &  Not Available \\	
	
	2015 &  \cite{garcia2017mongel}  & MoNGEL & No & No & None  &MkNN, OLM, OSDL  & \cite{MongelCode} in Java  \\

	2015 & \cite{li2015regularized} & RMC-FSVM & No & No & FSVM, SVM  & None  & Not Available \\
	
	2016 & 	\cite{gonzalez2016managing} & Monot. AdaBoost & No & No  & None & MID  & Not Available\\
	
	2016 & 	 \cite{Brookhouse2016} & AntMiner+, & No, No & 
	No, No  & ZeroR & OLM  & Not Available \\
	& 	                       & cAnt-Miner$_{PB}$+MC & &  &   &   & Not Available \\

	2016 &	\cite{garcia16Hiperrectangle}  & EHSMC-CHC & No & No & None & MkNN, OLM, OSDL, MID & Not Available \\ 
	
	2016 & \cite{chen2016xgboost}         & XGBoost  & No  & Yes  & pGBRT, Spark MLLib, H2O  & None & \cite{chen2016xgboost} in GitHub \\

	2016 & \cite{Bartley16PMSVM} &   PM-SVM & No & No & SVM & MC-SVM & \cite{Bartley16PMSVM} in GitHub\\
	2016 & \cite{Bartley16PMRF} &   PM-RF  & No & No & Random Forest  & MC-SVM & \cite{Bartley16PMRF} in GitHub\\

	2017 &	\cite{zhu2017monotonic} 	 & MCELM & No & Yes & CART, Rank Tree, ELM & OLM, OSDL, REMT & Not Available \\

	2017 & \cite{Verbeke17RULEM} & RULEM & No& Yes & Ripper, C4.5  & AntMiner+ &  Not Available\\
	
	2017 & \cite{alcala2017evolutionary} & MFARC-HD,  & No, No  & No, No & WM &  OSDL, M$k$NN, C4.5-MID,  &  Not Available\\
	&     & FS$_{MOGFS^{e}}$+T$_{UN^e}$ &  & &   & OLM, EHSMC-CHC, RF-MID & Not Available\\
	
	2018 & \cite{Bartley18MonoBoost} &  MonoBoost  & No & No & $k$NN & None & \cite{Bartley18MonoBoost} in GitHub\\
	
	
	
	\hline
\end{tabular}

	\end{table}
	
\end{landscape}

\subsection{Monotonic classifiers}
\label{subsec:MonotClassif}

\subsubsection{Instance based classifiers}
\label{subsec:InstBasClas}

\begin{itemize}
	\item Ordered Learning Model (\textbf{OLM} \cite{ben-david92,ben-david89}).
	The classification of new objects is done by the following function:
	\begin{equation}
	f_\mathrm{OLM}(\mathbf{x}) = \max \{  Y(\mathbf{x}_i): \mathbf{x}_i \in D, \mathbf{x}_i \preceq \mathbf{x}  \}.
	\end{equation}
	
	If there is no object from $D$ which is dominated by $\mathbf{x}$, then a class label is assigned by a nearest neighbor rule. $D$ is chosen to be consistent and not to contain redundant examples. An object $\mathbf{x}_i$ is redundant in $D$ if there is another object $\mathbf{x}_j$ such that $\mathbf{x}_i \succeq  \mathbf{x}_j$ and $Y(\mathbf{x}_i) = Y(\mathbf{x}_j)$.
	
	\item Ordered Stochastic Dominance Learner (\textbf{OSDL} \cite{lievens08,lievens10}). For each sample $\mathbf{x}_i$, OSDL computes two mapping functions: one that is based on the examples that are stochastically dominated by $\mathbf{x}_i$ with the maximum label (of that subset), and the second is based on the examples that cover (i.e., dominate) $\mathbf{x}_i$, with the smallest label. Later, an interpolation between the two class values (based on their position) is returned as a class.

	\item Monotonic k-Nearest Neighbor (\textbf{MkNN} \cite{duivesteijn08mKNN}). This classifier is an adaptation of the well-known nearest neighbor classifier, considering a full monotone data set. Starting from the original nearest neighbor rule, the class label assigned to a new data point $\mathbf{x}_0$ must lie in the interval $[y_\mathrm{min},y_\mathrm{max}]$, where:
	\begin{equation}
	y_\mathrm{min} = \max  \{ Y(\mathbf{x})|(\mathbf{x},Y(\mathbf{x})) \in D \wedge \mathbf{x}\preceq \mathbf{x}_0 \},
	\end{equation}
	and:
	\begin{equation}
	y_\mathrm{max} = \min \{ Y(\mathbf{x})|(\mathbf{x},Y(\mathbf{x})) \in D \wedge \mathbf{x}_0 \preceq \mathbf{x}  \}.
	\end{equation}

	\item \textbf{MOCA} (\cite{barile2008nonparametric}). MOCA is a nonparametric monotone
		classification algorithm that attempts to minimize the
		mean absolute prediction error for classification problems
		with ordered class labels. Firstly, the algorithm obtains a monotone classifier considering only training data. In the test phase, a simple interpolation scheme is applied.

\end{itemize}

\subsubsection{Decision trees and classification rules}
\label{subsec:DecTreesClasRules}

\begin{itemize}
	\item Monotonic Induction of Decision trees (\textbf{MID} \cite{ben-david95}). Ben-David introduces a measure of non-monotonicity in the	classical classification decision tree ID3 algorithm \cite{quinlan1986induction}. This measure was denoted as total-ambiguity-score. To calculate it, a non-monotonicity $b\times b$ matrix $M$ must be constructed, related to a tree containing $b$ branches. Each value $m_{ij}$ is 1 if the branches $i$ and $j$ are non-monotone, and 0 if they are.

	\item Positive Decision Tree, Quasi-Positive Decision Tree (\textbf{P-DT, QP-DT}  \cite{MAKINO1999}). 		
	In these algorithms the splitting rule considers to separate the points  that have the right child-node larger  than the left child-node (in the sense of the target variable).  
	The algorithm adds samples to the nodes in such way that the resulting tree is monotone.	
	This algorithm requires as precondition to be applied on strictly monotone binary data sets, containing only two classes.
	
	\item Monotonic Decision Tree (\textbf{MDT} \cite{lee2003monotonic}).	
	The authors proposed an induction approach to generate  monotonic decision trees from sets of examples which may not be monotonic or consistent. The algorithm constructs the tree using a set of ordinal labels which are not the same as the original ones. A mapping process can be used to relabel them into the originals.

	\item Isotonic Classification Tree (\textbf{ICT} \cite{van2009isotonic}).	
	This approach adjusts the probability estimated in the leaf nodes in case of a monotonicity violation. The idea is that, considering the monotonicity constraint, the sum of the absolute prediction errors on the training sample should be minimized. In addition, this 
	algorithm  can also handle problems where some, but not all, attributes have a monotonic relation with respect to the response.
	
	\item Rank Entropy based Monotonic decision Trees (\textbf{REMT} \cite{hu2012rank}).
	This algorithm introduces a metric called rank entropy as a robust measure of feature quality. It is used to compute the uncertainty, reflecting the ordinal structures in monotonic classification. The construction of the decision tree is based on this measure.

	\item \textbf{RDMT(H)} (\cite{marsala2015rank}).
	Marsala and Petturiti presented a tree classifier parametrized by a discrimination measure $H$, which is considered for splitting, together with other three pre-pruning parameters. RDMT(H) guarantees a weak form of monotonicity for the resulting tree when the data set is monotone consistent and \textit{H} refers to any rank discrimination measure. The authors adapted different measures to monotonic classification.
	
	\item \textbf{MGain} (\cite{zhang2015induction}).	
	MGain introduces the index of the monotonic consistency of a cut point with respect to a data set. When non-monotonic data appear in the training set, the index of monotonic consistency selects the best cut point. If the initial data set is totally monotonic, the results obtained are similar to those using C4.5 \cite{quinlan2014c4}.
	
	\item \textbf{AntMiner+, cAnt-Miner$_{PB}$+MC} (\cite{Brookhouse2016}).	
	This algorithm is an extension of an existing ant colony optimization based classification rule learner, able to create lists of monotonic classification rules. It considers an improved sequential covering strategy to search for the best list of classification rules.

	\item Rule Learning of ordinal classification with Monotonicity constraints (\textbf{RULEM}  \cite{Verbeke17RULEM}).	
	The authors present a technique to induce monotonic ordinal rule based classification models, which can be applied in combination with any rule or tree induction technique in a post processing step. They also introduce two metrics to evaluate the plausibility of the ordinal classification models obtained.

	\item \textbf{ MFARC-HD }(\cite{alcala2017evolutionary}).	
	In this case, different mechanisms based on monotonicity indexes are coupled with a popular and competitive classification evolutionary fuzzy system: FARC-HD. In addition, the proposal is able to handle any kind of classification data set without a preprocessing step.	
	
	\item \textbf{FS$_{MOGFS^{e}}$+T$_{UN^e}$} (\cite{alcala2017evolutionary}). The proposed method consists of two separated stages for learning and subsequent  tuning. The first stage is based on an improved multi-objective evolutionary algorithm designed to select the relevant features while learning the appropriate granularities of the membership functions. In the second stage, an evolutionary post-process is applied to the knowledge base obtained.

\end{itemize}

\subsubsection{Ensembles}
\label{subsec:multiclassifiers}

\begin{enumerate}
	\item Boosting

	\begin{itemize}
		
		\item \textbf{LPRules} (\cite{kotlowski09}).	
		This algorithm is based on a statistical analysis of the problem, trying to relate monotonicity constraints to the constraints imposed on the probability distribution. First, LPRules decomposes the problem into a sequence of binary subproblems. Then, the data for each subproblem is monotonized using a non-parametric approach by means of the class of all monotone functions.  In the last step, a rule ensemble is generated using the LPBoost method to avoid errors in the monotonized data.

		\item MOnotone Rule Ensembles (\textbf{MORE} \cite{dembczynski09}). 			
		MORE uses forward a stage-wise additive modeling scheme for generating an ensemble of decision rules for binary problems. An advantage of this method, as the authors indicate, is its comprehensibility and consistence.

		\item Monotonic Random Forest (\textbf{MonRF} \cite{Gonzalez2015}). 			
		The method is an adaptation of Random Forest \cite{breiman01} for classification with monotonicity constraints, including the rate of monotonicity as a parameter to be randomized during the growth of the trees. An ensemble pruning mechanism based on the monotonicity index of each tree is used to select the subset of the most monotonic decision trees which constitute the forest.

		\item Variable Consistency Dominance-based Rough Set Approach Ordinal Random Forest (\textbf{VC-DRSA ORF} \cite{wang2015ordinal}). 			
		The authors propose an Ordinal Random Forest based on the variable consistence dominance rough set approach. The ordinal random forest algorithm is implemented using Hadoop \cite{triguero15}.
		
		\item \textbf{Monotonic Adaboost} (\cite{gonzalez2016managing}).			
		In this case, decision trees are combined on an Adaboost scheme \cite{freund1995desicion}, considering a simple ensemble pruning method based on the degree of monotonicity. The objective in this algorithm is to offer a good trade-off between accurate predictive performance and the construction of monotonic models. 
		
		\item \textbf{XGBoost} (\cite{chen2016xgboost}).
		It is a open source library which provides the gradient boosting framework, and from  its 0.71 version, it supports monotonic constraints.

		\item Partially Monotone Random Forest (\textbf{PM-RF} \cite{Bartley16PMRF}).		
		By creating a novel re-weighting scheme, PM-RF is an effective partially monotone approach that was particularly good at retaining accuracy while correcting highly non monotone datasets with many classes, albeit only achieving monotonicity locally.
		
	\end{itemize}
	
	Absent from the literature there exist two publicly available and open source libraries: Arborist (\cite{Rborist})  and GBM (\cite{GBM}). Both are R packages that allow for monotone features by na\"{i}vely constraining each branch split (in each tree) to prohibit non monotonce splits.

	\item Bagging
	\begin{itemize}
		\item Variable Consistency Bagging (\textbf{VC-bagging} \cite{Blaszczynski2010}). 			
		For this proposal, the data set is
		structured using the Variable Consistency Dominance-based Rough Set
		Approach (VC-DRSA). A variable consistency bagging scheme is used to produce bootstrap samples that promotes classification examples with relatively high values of consistency measures.
		
		\item Fusing Rank Entropy based Monotonic decision Trees (\textbf{FREMT}  \cite{qian2015fusing,xu17FusingDecisTrees}). 			
		This method fuses decision trees taking into account attribute reduction and a fusing principle.
		The authors propose an attribute reduction
		approach with rank-preservation for learning base classifiers, which can effectively avoid overfitting and improve classification 	performance. In a second step, the authors establish a fusing principle considering the maximal probability through combining the base classifiers.

	\end{itemize}
	
\end{enumerate}

\subsubsection{Neural networks} \label{subsec:NeuralNetworks}

\begin{itemize}

	\item \textbf{Monotonic networks} (\cite{sill1997monotonic}).
	Monotonic networks implements a piecewise-linear surface by taking maximum and minimum operations on groups of hyperplanes. Monotonicity constraints are enforced by constraining the sign of the hyperplane weight.

	\item Monotonic Multi-Layer Perceptron (\textbf{MonMLP} \cite{lang2005monotonic}).
	This algorithm satisfies the requirements of monotonicity for one or more inputs by constraining the sign of the weights of the multi-layer perceptron network. The performance of MonMLP  does not depend on the quality of the training data because it is imposed in its structure.

	\item Monotonic Partial Neural Network MIN-MAX (\textbf{MPNN MIN-MAX} \cite{daniels10}).
	In this paper, the authors clarify some of the theoretical results on	
	monotone neural networks with positive weights, which sometimes produce misunderstood in the neural network literature. In addition, they generalize 
	of the so-called MIN-MAX networks to the case of partially monotone problems.

	\item Monotonic Classification Extreme Learning Machine (\textbf{MCELM} \cite{zhu2017monotonic}).
	MCELM is a generalization of extreme learning machine for monotonic classification data sets. The proposal involves a quadratic programing problem in which the monotonicity relationships are considered as constraints and the training errors as the objective to be minimized.

\end{itemize}

\subsubsection{Support Vector Machines}
\label{subsec:Statistical}

\begin{itemize}
	
	\item Monotonicity Constrained Support Vector Machine (\textbf{MC-SVM} \cite{chen2014credit,pelckmans2005primal}).
	MC-SVM is a rating model based on a support vector machine including monotonicity constraints in the optimization problem. The model is applied to credit rating, and the constraints are derived from the prior knowledge of financial experts.
	
	\item Regularized Monotonic Fuzzy Support Vector Machine (\textbf{RMC-FSVM} \cite{li2015regularized}).	
	This method applies the Tikhonov regularization \cite{tikhonov1977solutions} to SVMs with monotonicity constraints in order to ensure that the solution is unique and bounded. In this way, the prior domain knowledge of monotonicity can be represented in the form of inequalities based on the partial order of the training data.

	\item Partially Monotone Support Vector Machine (\textbf{PM-SVM} \cite{Bartley16PMSVM}).
	 PM-SVM differs from the MC-SVM by proposing a new constraint 	generation technique designed to more efficiently achieve monotonicity.

\end{itemize}

\subsubsection{Hybrid}
\label{subsec:Hybrid}

\begin{itemize}
	
	\item Monotonic Nested Generalized Exemplar Learning (\textbf{MoNGEL} \cite{garcia2017mongel}).	
	MoNGEL combines instance-based and rule learning. 
	The instances are converted to zero-dimensional rules, formed by a single point, obtaining an initial set of rules. As a second step, the method searches for that comparable rule of the same class with the minimum distance with respect to each rule, in order to iteratively generalize it. In the last step, the minimum number of rules which are non monotonic between them are removed.

	\item Evolutionary Hyperrectangle Selection for Monotonic Classification (\textbf{EHSMC-CHC} \cite{garcia16Hiperrectangle}).	
	After building a set of hyperrectangles from the training data set, a selection of them through evolutionary algorithms is applied. In a preliminary stage, an initial set of hyperrectangles are generated by using a heuristic based on the training data, and then a
	selection process is carried out, focused on maximizing the performance considering several objectives, such as accuracy, coverage of examples and reduction of the monotonicity violations of the model with the lowest possible number of hyperrectangles.

	\item \textbf{MonoBoost} (\cite{Bartley18MonoBoost}). 
	 Inspired by  instance based classifiers, MonoBoost is a framework for monotone additive rule ensembles where partial monotonocity appears. The algorithm ensures perfect partial monotonicity with reasonable performance. 
	
\end{itemize}

\subsubsection{Fuzzy Integrals}
\label{subsec:FuzzyInteg}

\begin{itemize}
	
	\item Heuristic Least Mean Square (\textbf{HLMS} \cite{grabisch1995new,grabisch1994classification}).
		HLMS aims to identifying the fuzzy measure taking advantage of the lattice structure of the coefficients. Thanks to this identification, the knowledge concerning the criteria can be obtained. 
	
	\item \textbf{Choquistic Regression} \cite{grabisch2003modelling,tehrani2012learning,tehrani2013ordinal}.
		The basic idea of choquistic regression is to replace the linear function of predictor variables,
		which is commonly used in logistic regression to model the log odds of the positive class,
		by the choquet integral \cite{grabisch1995fuzzy}.
	
\end{itemize}

\subsection{Monotonic Data Preprocessing}

Other group of methods in monotonic classification area are focused on applying preprocessing techniques to improve the performance of monotonic classification algorithms \cite{garcia2015data}. So far the literature proposals follow four paths:

\begin{enumerate}

	\item Relabeling. These methods aim at changing the class label of the instances which produce monotonicity violations to generate fully monotone data sets, which are required for many monotonic classifiers.

	\begin{itemize}
		
		
		\item \textbf{Dykstra Relabel} (\cite{dykstra1999nonparametric}). These authors propose a monotone relabeling based on isotonic regression, able to minimize absolute error or squared error. The algorithm is optimal optimizing those loss functions (absolute or squared error) but it does not guarantee the minimum number of label changes due to it is not the key objective.
		
		\item \textbf{Daniels-Velikova Greedy Relabel} (\cite{daniels2003derivation,feelders2006two}). 
		This is a greedy algorithm to relabel the non-monotone examples one at a time. At each step, it searches for the instance and the new label to
		maximize the increase in monotonicity of the data set. Although, at each step, it is able to maximize the jump towards complete monotonicity, the algorithm relabels more examples than is needed. This relabel method  does not guarantee an optimal solution.

		\item \textbf{Optimal Flow Network Relabel} (\cite{feelders2006two,rademaker2009loss,duivesteijn08mKNN}). This method is based on finding a maximum weight independent set in the monotonicity violation graph. Relabeling the complement of the maximum weight independent set results in a monotone data set with as few label changes as possible. This method is optimal, producing the minimal number of label changes.

		\item \textbf{Feelders Relabel} (\cite{feelders2010monotone,stegeman2011generating,feelders2016exploiting}). This algorithm faces the problem of relabeling with minimal empirical loss as a convex cost closure problem. Feelders relabel results in an optimal solution.

		\item \textbf{Single-pass Optimal Ordinal Relabel} (\cite{Rademaker2012}). In this case, the idea is to exploit the properties of a minimum flow network and identify pleasing properties of some maximum cuts. As expected by its name, this is an optimal relabeling algorithm.
		
		\item \textbf{Naive Relabel} (\cite{pijls2014repairing}). This algorithm is a building block of the two following ones, using a greedy scheme. The method does not guarantees an optimal solution.
		
		\item \textbf{Border Relabel} (\cite{pijls2014repairing}). This is a fast alternative to the greedy algorithm mentioned above, being more specific by minimizing the deviations between the new labels and the original ones. This case is similar to the previous one, being not optimal.
		
		\item \textbf{Antichain Relabel} (\cite{pijls2014repairing}). It is based in the previous one, minimizing the total number of relabellings. It leads to optimal solutions.
		
	\end{itemize}

	\item Feature Selection \cite{kotsiantis2011feature}. The objective of these methods is to improve the predictive capacity of the monotonic classifiers by selecting the most relevant characteristics. 
	
	\begin{itemize}
		\item \textbf{O-ReliefF, O-Simba} (\cite{hu2012large})
		
		The authors introduce margin-based feature selection algorithms for monotonic classification by incorporating the monotonicity constraints into the ordinal task. Relief and Simba methods are extended to the context of ordinal classification.
		
		\item min-Redundancy Max-Relevance (\textbf{mRMR} \cite{hu2012feature,pan2014feature,pan2016improved})
		
		The algorithm mRMR integrates the rank mutual information metric with the search strategy of min-redundancy and max-relevance, creating an effective algorithm for monotonic feature selection.
		
		\item Non-Monotonic feature selection via Multiple Kernel Learning (\textbf{NMMKL} \cite{yang2015budget}). Yang et al. propose a non-monotonic feature selection method that alleviates monotonic violations by computing the scores for individual features that depend on the number of selected features. 
		
	\end{itemize}
	
	\item Instance Selection \cite{cano2003using,garcia2012prototype}. The idea behind these algorithms is to improve the performance of monotonic classifiers by selecting the most useful instances to be used as training set, using instance-based heuristics.
	
	\begin{itemize}
		\item Monotonic Iterative Prototype Selection (\textbf{MONIPS} \cite{cano2017prototype})
		
		MONIPS follows an iterative scheme in which it determines the most representative instances which keep or improve the prediction capabilities of the MkNN algorithm. It follows an instance removal process based on the improvement of the MkNN performance.
		
	\end{itemize}
	
	\item Training Set Selection \cite{cano2007evolutionary}. This set of algorithms has the same objective as the previous ones, except that the heuristic followed must be generic in such a way that the selected set is the one that reports the highest performance regardless of the classifier that is used later on it.
	
	\begin{itemize}
		\item Monotonic Training Set Selection (\textbf{MonTSS} \cite{cano2017training})
		
		MonTSS incorporates
			proper measurements to identify and select the most suitable instances in
			the training set to enhance both the accuracy and the monotonic nature of
			the models produced by different classifiers.

	\end{itemize}
	
\end{enumerate}

\section{Quality metrics used in monotonic classification}
\label{subsec:qualityindexes}

\begin{table}
	\caption{Metrics considered in the reviewed monotonic classification methods.}
	\label{tab:metrics}       
	\tiny
	\begin{tabular}{lcc}
		\hline\noalign{\smallskip}
		& Predictive assessment  & Monotonicity   \\
		Abbr. name           &  metrics          &     fulfillment  metrics             \\     
		\noalign{\smallskip}\hline\noalign{\smallskip}
		
		OLM &  MSE & None   \\		
		
		MID &  MSE, MAE & NMI   \\	
		
		HLMS  & Accuracy & None \\
		Monotonic Networks & Error Rate & None \\
		
		P-DT, QP-DT &  Error Rate & None\\

		MORE & MAE  & None  \\
		
		MDT & Accuracy  & $\gamma_1$, $\gamma_2$ \\

		MonMLP & None & None \\
		
		OSDL & None   & None \\

		MkNN & Error Rate & None  \\		
		
		MOCA & MAE & None\\
		
		ICT & MAE & None \\
		
		LPRules & MAE &  None \\
		
		MPNN MIN-MAX & MSE, Error Rate  & None \\
		
		VC-bagging & MAE & None \\		
		
		REMT & MAE  & None \\
		
		Choquistic Regression & Accuracy, AUC & None \\
		
		MC-SVM & Accurary, Recall, PPV,   & FOM  \\
		&    NPV, F-Measure., $\kappa$ coefficient            & \\
		MGain & Accuracy  & None   \\
		FREMT & Accuracy, MAE &  None \\	
		MonRF & Accuracy, MAE & NMI\\	
		
		VC-DRSA ORF & None  & None \\	
		
		RDMT(H) & Accuracy, $\kappa$ coefficient, MAE & NMI \\	
		
		MoNGEL &  Accuracy, MAE & NMI\\
		
		RMC-FSVM & Accuracy, Recall,    & None \\
		
		&  PPV, F-Measure     & \\

		Monot. AdaBoost  & Accuracy, MAE & NMI\\
		
		AntMiner+, cAnt-Miner$_{PB+MC}$& Accuracy  & None  \\

		EHSMC-CHC & Accuracy, MAE, MAcc, MMAE & NMI  \\ 
		
		XGBoost & AUC & None\\
		
		PM-SVM &  Accuracy, $\kappa$ coefficient & MCC \\
		PM-RF & & \\
		
		MCELM & MAE & None\\

		RULEM &  Accuracy, MAE, MSE & None\\
		MFARC-HD, FS$_{MOGFS^{e}}$+T$_{UN^e}$ &  MAE, MMAE  & NMI \\
		MonoBoost & F-Measure, $\kappa$ coefficient, Recall, Accuracy  & None\\
		\noalign{\smallskip}\hline
	\end{tabular}
\end{table}

This section analyzes and summarizes the evaluation measures used in all the experimental studies present in the specialized literature. They evaluate two different aspects: precision and monotonicity. In Table \ref{tab:metrics}, we present, for each monotonic classification method, the measures used both for predictive assessment and for monotonicity fulfillment. The description of each metric is included below.

\subsection{Predictive assessment metrics}

In order to define the metrics considered to evaluate the predictive performance of a classifier, we introduce the following notation:
\begin{itemize}
	\item True Positives (TP): number of instances with positive outcomes that are correctly classified.
	\item False Positives (FP): number of instances with positive outcomes that are incorrectly classified.
	\item True Negative (TN): number of instances with negative outcomes that are correctly classified.
	\item False Negative (FN): number of instances with negative outcomes that are incorrectly classified.
\end{itemize}

The first set of predictive measures included are applied in binary classification, and they are listed below: 

\begin{itemize}

	\item Accuracy (\cite{chen2014credit}):
	\begin{equation}
	\text{Accuracy} = \frac{TP+TN}{TP+FP+TN+FN},
	\end{equation}
	
	representing the predictive ability according to the proportion of the tested data correctly classified. 
	
	\item Error rate (\cite{chen2014credit}):
	
	\begin{equation}
	\text{Error Rate} = \frac{FP+FN}{TP+FP+TN+FN}.			
	\end{equation}
	
	This is the opposite case to the previous one, evaluating the proportion of the tested data incorrectly classified.

	\item Recall (\cite{chen2014credit}):
	\begin{equation}
	\text{Recall} = \frac{TP}{TP+FN}.
	\end{equation}
	
	Recall (also called sensitivity) is a measure of the proportion of actual positives that are correctly classified.

	\item Positive predictive value (PPV \cite{chen2014credit}):
	\begin{equation}
	\text{PPV} = \frac{TP}{TP+FP},
	\end{equation}		
	which is the proportion of test instances with positive predictive outcomes that are correctly predicted. PPV (also known as precision) represents the probability that a positive test reflects the underlying condition being tested for.
	
	\item Negative predictive value (NPV \cite{chen2014credit}):
	\begin{equation}
	\text{NPV} = \frac{TN}{TN+FN},
	\end{equation}		
	which is the proportion of test instances with negative predictive outcomes that are correctly predicted.

	\item F-Measure (\cite{chen2014credit}):
	\begin{equation}
	\text{F-Measure} = \frac{2 \cdot PPV \cdot Recall}{PPV+Recall}.
	\end{equation}		
	This metric is the harmonic mean of precision and recall.
	
	\item The $\kappa$ coefficient (\cite{chen2014credit}) represents the agreement between the classifier and the data labels, and it is computed as follows:
	\begin{equation}
	\kappa \text{ coefficient} = \frac{P_a -P_e}{1-P_e},
	\end{equation}		
	where $P_e$ is the hypothetical probability of chance agreement and $P_a$ is the relative observed agreement between the classifier and the data. They are computed as follows:
	\begin{align}
	\text{P}_e = \frac{(TP+FP) \cdot (TP+FN) + (TN+FP) \cdot (TN+FN)}{ (TP+TN+FP+FN)^2},\\
	\text{P}_a = \frac{TP+TN}{TP+TN+FP+FN}.
	\end{align}		
	
	\item Area Under Curve (AUC): To combine the Recall and the false positive rate ($\frac{FP}{FP+TN}$)  into one single metric, we first compute the two former metrics with many different threshold (for example 0.00, 0.01, 0.02, ... ,1.00) for the logistic regression, then plot them on a single graph, with the false positive rate values on the abscissa and the Recall values on the ordinate. The resulting curve is called ROC curve, and the metric we consider is the AUC of this curve.
	
\end{itemize}

The second set set of predictive measures are applied in multiclass classification problems, and they are listed below:

\begin{itemize}
	
	\item Mean Squared Error (MSE \cite{Verbeke17RULEM}) is calculated as:
	\begin{equation}
	\text{MSE} = \frac{1}{n}\sum_{i=1}^{n}(y'_i-y_i)^2,
	\end{equation}
	where $n$ is the number of observations in the evaluated data set, $y'_i$ the estimated class label for observation $i$ and $y_i$ the true class label (both represented as integer values based on their position in the ordinal scale). It measures the average of the squares of errors.
	
	\item Mean Absolute Error (MAE \cite{Verbeke17RULEM}) is defined as:
	\begin{equation}
	\text{MAE} = \frac{1}{n}\sum_{i=1}^{n}|y'_i-y_i|.
	\end{equation}
	MAE is a measure of how close predictions are to the outcomes.
	
	\item Monotonic Accuracy (MAcc \cite{garcia16Hiperrectangle}), computed as standard Accuracy, but only considering those examples
	that completely fulfill the monotonicity constraints in the test set. In other words, non-monotonic
	examples do not take part in the calculation of MAcc.
	
	\item Monotonic Mean Absolute Error (MMAE \cite{garcia16Hiperrectangle}), calculated  as standard MAE, but only considering those examples that completely fulfill the monotonicity constraints in the test set.
	
	\item Monotonicity Compliance (MCC \cite{Bartley16PMSVM}), defined as the proportion of the input space where the requested monotonicity constraints are not violated, weighted by the joint probability distribution of the input space. This metric has been proposed to be applied when partial monotonicity is present.
	
\end{itemize}

\subsection{Monotonicity fulfillment metrics}

In this case, the interest is to evaluate the rate of monotonicity provided by either the predictions obtained or the model built.

Let $\mathbf{x}$ be an example from the data set $D$.  $NClash(\mathbf{x})$ is the number of examples from $D$ that do not meet the monotonicity restrictions with respect to $\mathbf{x}$, and $n$ is the number of instances in $D$. $NMonot(\mathbf{x})$ is the number of examples from $D$ that  meet the monotonicity restrictions with respect to $\mathbf{x}$.

\begin{itemize}

	\item The Non-Monotonic Index(\cite{ben-david95,Daniels06NMI1}) is defined as the number of clash-pairs divided by the total number of pairs of examples in the data set:	
	\begin{equation}
	NMI= \dfrac{1}{n(n-1)}\sum_{\mathbf{x} \in D} NClash(\mathbf{x})
	\end{equation}
	
	\item $\gamma_1$ (\cite{lee2003monotonic}), assessed as:
	\begin{eqnarray}
	\gamma_1 = \frac{S_{+}-S_{-}}{S_{+}+S_{-}},	 \\
	S_{-} = \sum_{\mathbf{x} \in D} NClash(\mathbf{x}),  \\
	S_{+} = \sum_{\mathbf{x} \in D} NMonot(\mathbf{x}), 
	\end{eqnarray}	
	where $S_{-}$ is the number of discordant pairs, and  $S_{+}$ is the number of concordant pairs.  $\gamma_1$  is the Goodman-Kruskal's $\gamma$ statistic (\cite{goodman79}).

	\item $\gamma_2$ (\cite{lee2003monotonic}):	
	\begin{eqnarray}
	\gamma_2 = \frac{S_{+}-S_{-}}{ \#P },
	\end{eqnarray}		
	where $\#P$ is the total number of pairs, i.e. $P=S_{+}+S_{-}+\#NCP$, $\#NCP$ standing for number of non-comparable pairs.
	
	\item Frequency of Monotonicity (FOM \cite{chen2014credit}):
	\begin{equation}
	\text{FOM} = \frac{S_{+}}{ \#P }.
	\end{equation}
	
	\item The Non-Monotonicity Index 2 (NMI2 \cite{milstein14}) is defined as the number of non-monotone examples divided by the total number of examples:
	\begin{equation}
	NMI2= \dfrac{1}{n}\sum_{\mathbf{x} \in D} Clash(\mathbf{x})
	\end{equation}
	where $Clash(\mathbf{x})$ = 1 if $\mathbf{x}$ clashes with at least one example in $D$, and 0 otherwise. If $Clash(\mathbf{x})=1$, $\mathbf{x}$ is called a non-monotone example. This metric was proposed in \cite{milstein14} but it has not been used in any study yet.
\end{itemize}

Table \ref{tab:metricsFreq} includes the number of times each metric was used in the different experimental studies. As can be observed, the most commonly used metrics for predictive purposes are Accuracy and MAE, whereas NMI is the most popular one for estimating the monotonicity fulfillment.

\begin{table}
	\centering
	\caption{Number of times each metric is used in monotonic classification literature.}
	\label{tab:metricsFreq}       
	\begin{tabular}{lcc}
		\hline\noalign{\smallskip}
		Metric & \# of times used \\
		\noalign{\smallskip}\hline\noalign{\smallskip}
		MAE & 17   \\
		Accuracy  & 19 \\ 	
		Error Rate & 5  \\	
		$\kappa$ coefficient  & 5\\			
		MSE &  4  \\

		Recall & 3 \\
		F-Measure  & 3 \\		
		PPV  & 2 \\

		MMAE &  2  \\	 				
		AUC & 2 \\	
		NPV & 1 \\	
		MAcc & 1   \\

		\hline
		
		NMI & 7 \\ 	
		MCC & 2 \\	 
		$\gamma_1$  &  1 \\
		$\gamma_2$ & 1 \\ 
		FOM & 1 \\

		NMI2 & 0 \\

		\noalign{\smallskip}\hline
	\end{tabular}
\end{table}

\section{Data sets used in monotonic classification}
\label{sec:datasets}

Next, we review monotonic classification papers to summarize which are the data sets considered in their experimental analysis.

The information about the most commonly used data sets (with at least 10 appearances in the literature) has been included in Table \ref{tab:ResumenUsualDataSets}, which summarizes their properties. For each data set, we can observe the number of examples (Ex.), attributes (Atts.), numerical attributes (Num.) and nominal attributes (Nom.), the number of classes (Cl.), the source where the data set can be found, and finally, the number of times included in the experimental analysis in the literature.

\begin{table} 
	{\normalsize
		\caption{Summary of the most used data sets used in the monotonic classifiers literature.}
		\label{tab:ResumenUsualDataSets}
	}	
	\centering
	\begin{tabular}{lccccccc}
		\hline	
		Data Set  & Ex.  &  Atts. & Num. & Nom. & Cl. & Source & \# of times used \\
		\hline		
		AutoMPG   &  392     &    7  & 7  & 0  & 10 & \cite{alcala11} & 16 \\			
		BostonHousing& 506 & 12  & 10 & 2 & 4 & \cite{UCI13} & 12 \\
		Car& 1728 & 6 & 0 & 6 & 4 & \cite{alcala11} & 17 \\
		ERA& 1000 & 4 & 4 & 0 & 9 & \cite{ben-david89} & 14 \\
		ESL& 488  & 4 & 4 & 0 & 9 & \cite{ben-david89} & 17 \\
		LEV& 1000 & 4 & 4 & 0 & 5 & \cite{ben-david89} & 13 \\
		MachineCPU& 209 & 6 & 6 & 0 & 4 & \cite{alcala11} & 15\\
		Pima& 768 & 8 & 8 & 0  & 2  &  \cite{alcala11} & 14 \\
		
		SWD& 1000 & 10 & 10 & 0 & 4 & \cite{ben-david89} & 14 \\
		\hline

	\end{tabular}

\end{table}

A brief description is now given for each of these data sets:
\begin{itemize}
	\item AutoMPG: the data concerns city-cycle fuel consumption in miles per gallon (Mpg).	
	\item BostonHousing: the data set concerns the housing values in suburbs of Boston.
	\item Car: this data set (Car Evaluation Database) was derived from a simple hierarchical decision model. The model evaluates cars according to six input attributes: buying, maint, doors, persons, lug\_boot, safety.
	\item ERA: this data set was originally gathered during an academic decision-making experiment aiming at determining which are the most important qualities of candidates for a certain type of jobs. 
	\item ESL: in this case, we find profiles of applicants for certain industrial jobs.  Expert psychologists of a recruiting company, based on psychometric test results and interviews with the candidates, determined the values of the input attributes. The output is an overall score corresponding to the degree of fitness of the candidate to this type of job.
	\item LEV: this data set contains examples of anonymous lecturer evaluations, taken at the end of MBA courses. Before receiving the final grades, students were asked to score their lecturers according to four attributes such as  oral skills and contribution to their professional/general knowledge. The single output was a total evaluation of the lecturer's performance.	
	\item Pima: this data set comes from the National Institute of Diabetes and Digestive and Kidney Diseases. Several constraints were placed on the selection of sample from a larger database. In particular, all patients here are females of Pima Indian heritage, which are, at least, 21 years old. The class label represents if the person has (or not) diabetes. 
	\item MachineCPU: this problem focuses on relative CPU performance data. The task is to approximate the published relative performance of the CPU. 
	\item SWD: it contains real-world assessments of qualified social workers regarding the risk of a group of children if they stay with their families at home.  This evaluation of risk assessment is often presented to judicial courts to help decide what is in the best interest of an alleged abused or neglected child.
\end{itemize}

Considering these data sets, Table \ref{tab:ResumenFeatUsualDataSets} includes the estimation of the possible monotonic relationship between each input feature and the class feature, using for this the RMI measure \cite{hu12}. This metric takes values in the range $[-1, 1]$, where $-1$ means that the relation is totally inverse (if the feature increases, the class decreases), and $1$ represents a completely direct relation (if the feature increases, the class increases). If the relation is direct (for instance, a value in the range $[0.1,1]$), we include a '+' in the cell. In the case of inverse relation (a value in the range $[-1,-0.1]$), the symbol used is '-', and, when the RMI value is in the range $[-0.1,0.1]$, we consider that there is no relation between the feature and the class (represented by a '='). The RMI value is given below each corresponding symbol. As can be checked in Table \ref{tab:ResumenFeatUsualDataSets}, most of the characteristics present a relation with the corresponding class, so that they are good candidate data sets to be used in future experimental studies.

\begin{table}
	{\normalsize
		\caption{RMI measure \cite{hu12} for all input features when considering the most popular monotonic classification data sets.}
		\label{tab:ResumenFeatUsualDataSets}
	}	
	\centering
	\begin{tabular}{lcccccccccccc}
		\hline	
		Data Set  & A1  &  A2 & A3 & A4 & A5 & A6 & A7  &  A8 & A9 & A10 & A11 & A12 \\
		\hline	
		AutoMPG & - & - & - & - & + & + & + &  &  &  &  & \\ 
		& -.5 & -.8 & -.8 & -.7 & .3 & .6 & .4 &  &  &  &  & \\ 
		BostonHousing & - & + & - & = & - & + & - & + & - & - & - & = \\
		& -.5 & .3 & -.4 & .0 & -.4 & .6 & -.4 & .2 & -.2 & -.4 & -.5 & .0 \\
		Car & + & + & + & + & + & + &  &  &  &  &  &  \\
		& 1. & 1. & 1. & 1. & 1. & 1. &  &  &  &  &  &  \\
		
		ERA & + & + & + & + &  &  &  &  &  &  &  &  \\
		& .3 & .4 & .2 & .2 &  &  &  &  &  &  &  &  \\
		
		ESL & + & + & + & + &  &  &  &  &  &  &  &  \\
		& .6 & .6 & .6 & .6 &  &  &  &  &  &  &  &  \\
		
		LEV & + & + & + & + &  &  &  &  &  &  &  &  \\
		& .2 & .4 & .2 & .2 &  &  &  &  &  &  &  &  \\
		
		MachineCPU & - & + & + & + & + & + &  &  &  &  &  &  \\
		& -.6 & .6 & .7 & .7 & .5 & .5 &  &  &  &  &  &  \\
		
		Pima & +  & +   & =  & =  & +  & +  & +  & +  &  &  &  &  \\
		& .2  & .3  & .0 & .0 & .2 & .2 & .2 & .2 &  &  &  &  \\

		SWD & + & + & +    & = & + & = & + & = & + & + &  & 	 \\	
		& .2 & .2 & .3 & .0 & .2 & .0 & .2 & .0 & .2 & .2 &  & 	 \\	
		\hline

	\end{tabular}

\end{table}

\section{Conclusions}
\label{sec:conclusions}

This paper is a systematical review of monotonic classification literature that could be used as a functional guide on the scope. Monotonic classification is an emerging area in the field of data mining. In recent years, the number of proposals in this area of knowledge has significantly increased, as shown in Figure \ref{fig:numProposals}. This fact justifies the necessity of proposing a taxonomy that classifies and discriminates all the methods proposed so far. The taxonomy designed can be used as a guide to:
\begin{itemize}
	\item Decide which kind of algorithm and model is interesting for a new monotonic  problem.
	\item Consider existing methods as reference to develop new proposals in this area of study.
	\item Compare any new proposal with the previous ones in the same family, to confirm whether or not it improves their performance and deserves to be considered.
\end{itemize}

Together with this taxonomy, we also analyze which methods are publicly available, their source code being available on line. In those cases, we also include where their implementation can be found.

Additionally, an analysis of the proposed and used quality metrics is carried out, considering predictive assessment and monotonicity fulfillment. We also highlight some measures, which are more frequently considered in this field, such as Accuracy, MAE and NMI.

Finally, a summary and description of all the data sets used is considered. We emphasize eight of them, which have been used in, at least, ten of the experimental evaluations reviewed in the papers. Their characteristics, availability and the monotonic relationships between input features and the class label are also reported.

\section*{Acknowledgement} 

This work has been supported by TIN2017-89517-P, by TIN2015-70308-REDT, by TIN2014-54583-C2-1-R and the Spanish ``Ministerio de
Econom{\'i}a y Competitividad" and by ``Fondo Europeo de
Desarrollo Regional" (FEDER) under Project TEC2015-69496-R.

\section*{References}

\bibliography{Bibliography}

\end{document}